\documentclass[11pt,a4paper]{article}
\usepackage{times,latexsym}
\usepackage{url}
\usepackage[T1]{fontenc}
\usepackage{graphicx}
\usepackage{amsmath}
\usepackage{amssymb}
\usepackage{booktabs}
\usepackage{lipsum}
\usepackage{graphicx}
\usepackage{bm}
\usepackage{xcolor}
\usepackage{stmaryrd}
\usepackage{multicol}
\usepackage{scalerel}
\usepackage{multirow}
\usepackage{nccmath}

\usepackage{color, colortbl}
\definecolor{Gray}{gray}{0.93}

\usepackage{array}
\newcolumntype{P}[1]{>{\raggedright\arraybackslash}p{#1}}
\newcolumntype{N}[1]{>{\raggedright\arraybackslash\columncolor{Gray}}p{#1}}

\usepackage[acceptedWithA]{tacl2018v2}

\usepackage{xspace,mfirstuc,tabulary}

\definecolor{knolcol}{rgb}{0.0,0.2,0.4}
\definecolor{humancol}{rgb}{0.0,0.2,0.4}
\definecolor{robotcol}{rgb}{0.0,0.0,0.0}
\definecolor{topiccol}{rgb}{0.0,0.0,0.0}

\newif\iftaclinstructions
\taclinstructionsfalse 
\iftaclinstructions
\renewcommand{\confidential}{}
\renewcommand{\anonsubtext}{(No author info supplied here, for consistency with
TACL-submission anonymization requirements)}
\newcommand{\instr}
\fi

\iftaclpubformat 

\else

\fi

\title{Modeling Global and Local Node Contexts \\ for Text Generation from Knowledge Graphs}

\author{Leonardo F. R. Ribeiro$^{\dag}$, Yue Zhang$^{\ddag}$, Claire Gardent$^{\S}$ and Iryna Gurevych$^{\dag}$ \vspace{1mm} \\
\rule{0pt}{2.5ex}
  $^{\dag}$Research Training Group AIPHES and UKP Lab, Technische Universit\"at Darmstadt\\
  $^{\ddag}$School of Engineering, Westlake University, $^{\S}$CNRS/LORIA, Nancy, France \\
 \texttt{ribeiro@aiphes.tu-darmstadt.de}, \texttt{yue.zhang@wias.org.cn} \\
 \texttt{claire.gardent@loria.fr}, \texttt{gurevych@ukp.informatik.tu-darmstadt.de}
}

\date{}

\begin{document}
\maketitle
\begin{abstract}
  Recent graph-to-text models generate text from graph-based data using either global or local aggregation to learn node representations. \emph{Global node encoding} allows explicit communication between two distant nodes, thereby neglecting graph topology as all nodes are directly connected. In contrast, \emph{local node encoding} considers the relations between neighbor nodes capturing the graph structure, but it can fail to capture long-range relations. In this work, we gather both encoding strategies, proposing novel neural models which encode an input graph combining both global and local node contexts, in order to learn better contextualized node embeddings. In our experiments, we demonstrate that our approaches lead to significant improvements on two graph-to-text datasets achieving BLEU scores of 18.01 on AGENDA dataset, and 63.69 on the WebNLG dataset for seen categories, outperforming state-of-the-art models by 3.7 and 3.1 points, respectively.\footnote{Code is available at \href{https://github.com/UKPLab/kg2text}{https://github.com/UKPLab/kg2text}}
  

\end{abstract}

\section{Introduction}





 \begin{figure}[t]
    \centering
    \includegraphics[width=.43\textwidth]{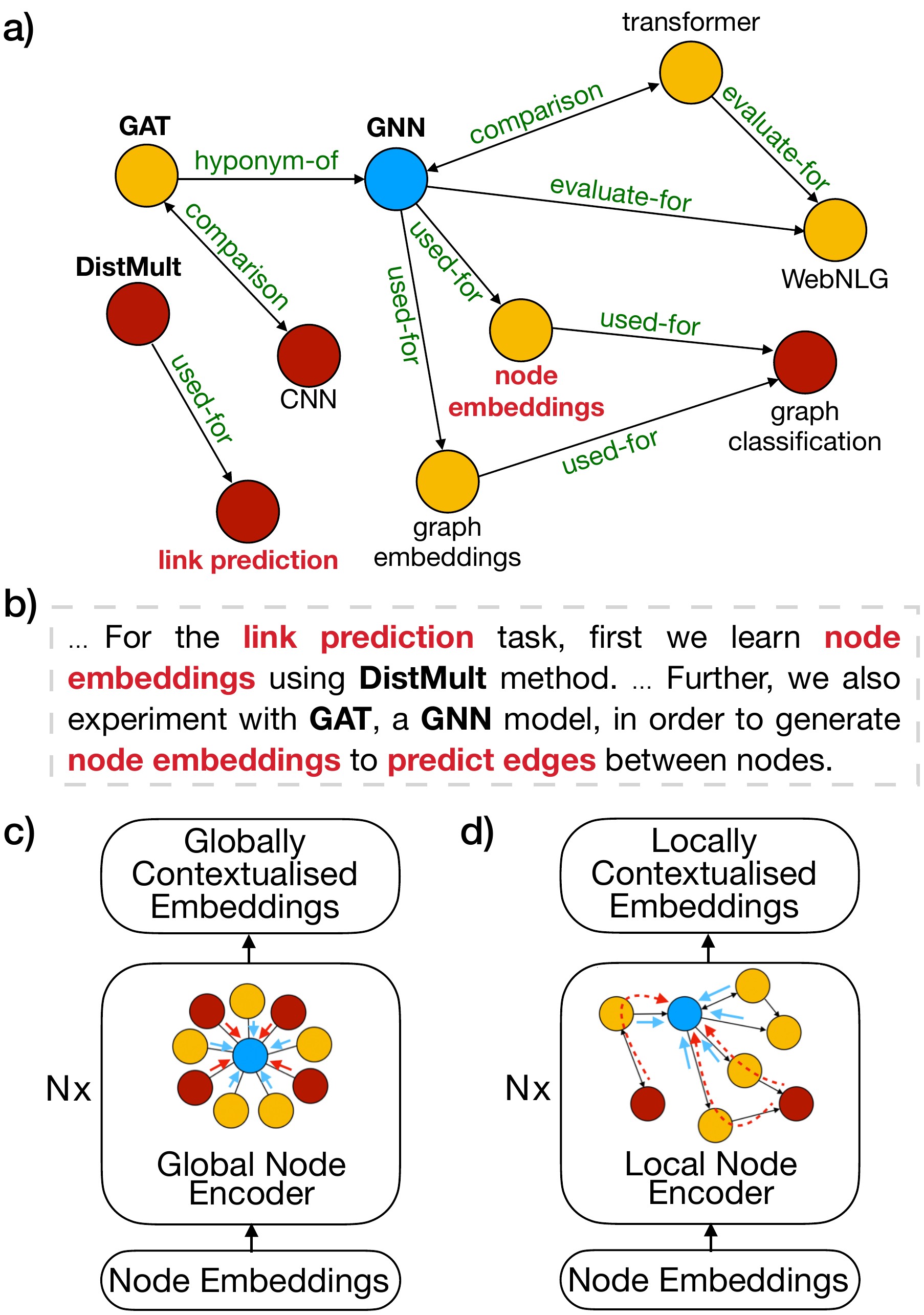}
    \caption{A graphical representation (a) of a scientific text (b). (c) A global encoder directly captures longer dependencies between any pair of nodes (blue and red arrows), but fails in capturing the graph structure. (d) A local encoder explicitly accesses information from the adjacent nodes (blue arrows) and implicitly captures distant information (dashed red arrows). }
    \label{fig:kg}
\end{figure}

Graph-to-text generation refers to the task of generating natural language text from input graph structures, which can be semantic representations \cite{konsas_17} or knowledge graphs (KG) \cite{gardent-etal-2017-webnlg, koncel-kedziorski-etal-2019-text}.
While most recent work \cite{song-etal-acl2018, ribeiro-etal-2019-enhancing, dcgcnforgraph2seq19guo} focuses on generating sentences, 
a more challenging and interesting scenario emerges when the goal is to generate 
multi-sentence texts. In this context, in addition to sentence generation, document planning needs to be handled: 
the input needs to be mapped into several sentences; sentences need to be ordered and connected using appropriate discourse markers; and inter-sentential anaphora and ellipsis may need to be generated to avoid repetition. In this paper, we focus on generating texts rather than sentences where the output are short texts \cite{gardent-etal-2017-webnlg} or paragraphs \cite{koncel-kedziorski-etal-2019-text}.

A key issue in neural graph-to-text generation is how to encode the input graphs. The basic idea is to incrementally compute node representations by aggregating structural context information.  
To this end, two main approaches have been proposed: (i) models based on \textit{local node aggregation}, usually built upon Graph Neural Networks (GNN) \cite{Kipf:2016tc, NIPS2017_6703} and (ii) models that leverage \textit{global node aggregation}. Systems that adopt global encoding strategies are typically based on Transformers \cite{NIPS2017_7181}, using self-attention to compute a node representation based on all nodes in the graph. This approach enjoys the advantage of a large node context range, but neglects the graph topology by effectively treating every node as being connected to all the others in the graph. In contrast, models based on local aggregation learn the representation of each node based on its adjacent nodes as defined in the input graph. This approach effectively exploits the graph topology, and the graph structure has a strong impact on the node representation~\cite{Xu2018RepresentationLO}. However, encoding relations between distant nodes can be challenging by requiring more graph encoding layers, which can also propagate noise \cite{li2018deeper}.

For example, Figure~\ref{fig:kg}a presents a KG, for which a corresponding text is shown in Figure~\ref{fig:kg}b. Note that there is a mismatch between how entities are connected in the graph and how their natural language descriptions are related in the text. Some entities syntactically related in the text are not connected in the graph. For instance, in the sentence "For \textit{the link prediction task}, first we learn \textit{node embeddings} using \textit{DistMult method}.", while the entity mentions are dependent of the same verb, in the graph, the \textit{node embeddings} node has no explicit connection with \textit{link prediction} and \textit{DistMult} nodes, which are in a different connected component. This example illustrates the importance of encoding distant information in the input graph. As shown in Figure~\ref{fig:kg}c, a global encoder is able to learn a node representation for \textit{node embeddings} which captures information from non-connected entities such as \textit{DistMult}. By modeling distant connections between all nodes, we allow for these missing links to be captured, as KGs are known to be highly incomplete \cite{45634, Schlichtkrull2018ModelingRD}. 

In contrast, the local strategy refines the node representation with richer neighborhood information, as nodes that share the same neighborhood exhibit a strong \textit{homophily}: two similar entities are much more likely to be connected than at random. Consequently, the local context enriches the node representation with local information from KG triples. For example, in Figure~\ref{fig:kg}a, \textit{GAT} reaches \textit{node embeddings} through the \textit{GNN}. This transitive relation can be captured by a local encoder, as shown in Figure~\ref{fig:kg}d. Capturing this form of relationship also can support text generation at the sentence level.


In this paper, we investigate novel graph-to-text architectures that combine both \textit{global} and \textit{local} node aggregations, gathering the benefits from both strategies. In particular, we propose a unified graph-to-text framework based on Graph Attention Networks (GAT) \cite{velickovic2018graph}. As part of this framework, we empirically compare two main architectures: a \emph{cascaded architecture} that performs global node aggregation before performing local node aggregation, and a \emph{parallel architecture} that performs global and local aggregations simultaneously. While the cascaded architecture allows the local encoder to leverage global encoding features, the parallel architecture allows more independent features to complement each other. To further consider fine-grained integration, we additionally consider \emph{layer-wise integration} of the global and local encoders. 

Extensive experiments show that our approaches consistently outperform recent models on two benchmarks for text generation from KGs. To the best of our knowledge, we are the first to consider integrating global and local context aggregation in graph-to-text generation, and the first to propose a unified GAT structure for combining global and local node contexts. 

\section{Related Work}

Early efforts for graph-to-text generation employ statistical methods \cite{flanigan-etal-2016-generation, pourdamghani-etal-2016-generating, song-etal-2017-amr}. Recently, several neural \mbox{graph-to-text} models have exhibited success by leveraging encoder mechanisms based on LSTMs, GNNs and Transformers.

 
 \paragraph{AMR-to-Text Generation.} Various neural models have been proposed to generate  sentences from Abstract Meaning Representation (AMR) graphs. \citet{konsas_17} provide the first neural approach for this task, by linearising the input graph as a sequence of nodes and edges. \citet{song-etal-acl2018} propose the graph recurrent network (GRN) to directly encode the AMR nodes, whereas \citet{beck-etal-2018-acl2018} develop a model based on gated GNNs. However, both approaches only employ local node aggregation strategies. \citet{damonte_naacl18} combine graph convolutional networks (GCN) and LSTMs in order to learn complementary node contexts. However, differently from Transformers and GNNs, LSTMs generate node representations that are influenced by the node order. \citet{ribeiro-etal-2019-enhancing} develop a model based on different GNNs which learns node representations which simultaneously encode a top-down and a bottom-up views of the AMR graphs, whereas \citet{dcgcnforgraph2seq19guo} leverage dense connectivity in GNNs. Recently, \citet{doi:10.116200297} propose a local graph encoder based on Transformers using separated attentions for incoming and outgoing neighbors. Recent methods \cite{zhu-etal-2019-modeling,cai-lam-2020-graph} also employ Transformers, but learn globalized node representations, modeling graph paths in order to capture structural relations. 

\paragraph{KG-to-Text Generation.} In this work, we focus on generating text from KGs. In comparison to AMRs, which are rooted and connected graphs, KGs do not have a defined topology, which may vary widely among different datasets, making the generation process more demanding. KGs are sparse structures that potentially contain a large number of relations. Moreover, we are typically interested in generating multi-sentence texts from KGs, which involves solving document planning issues \cite{konstas-lapata-2013-inducing}. 

Recent neural approaches for KG-to-text generation simply linearise the KG triples thereby loosing graph structure information.  For instance, \citet{colin-gardent-2018-generating}, \citet{moryossef-etal-2019-step} and Adapt \cite{gardent-etal-2017-webnlg} employ LSTM/GRU to encode WebNLG graphs. \citet{castro-ferreira-etal-2019-neural} systematically compare pipeline and end-to-end models for text generation from WebNLG graphs.
\citet{trisedya-etal-2018-gtr} develop a graph encoder based on LSTMs that captures relationships within and between triples.
Previous work has also studied how to explicitly encode the graph structure using GNNs or Transformers. \citet{marcheggiani-icnl18} propose an encoder based on GCNs, which consider explicitly local node contexts, and show superior performance compared to LSTMs. Recently, \citet{koncel-kedziorski-etal-2019-text} propose a Transformer-based approach which computes the node representations by attending over node neighborhoods following a self-attention strategy. In contrast, our models focus on distinct global and local message passing mechanisms, capturing complementary graph contexts. 

\paragraph{Integrating Global Information.} There has been recent work that attempts to integrate global context in order to learn better node representations in graph-to-text generation. To this end, existing methods employ an artificial global node for message exchange with the other nodes. This strategy can be regarded as extending the graph structure but using similar message passing mechanisms. In particular, \citet{koncel-kedziorski-etal-2019-text} add a global node to the graph and use its representation to initialize the decoder. Recently, \citet{dcgcnforgraph2seq19guo} and \citet{cai-lam-2020-graph} also employ an artificial global node with direct edges to all other nodes to allow global message exchange for AMR-to-text generation. Similarly, \citet{zhang-etal-2018-sentence} use a global node to a GRN model for sentence representation. Different from the above methods, we consider integrating global and local contexts at the {\it node} level, rather than the {\it graph} level, by investigating {\it model} alternatives rather than {\it graph structure} changes. In addition, we integrate GAT and Transformer architectures into a unified global-local model.



\section{Graph-to-Text Model}
\label{sec:background}


This section first describes (i) the graph transformation adopted to create a relational graph from the input (Section~\ref{sec:graph_preparation}), and (ii) the graph encoders of our framework based on Graph Attention Networks (GAT) \cite{velickovic2018graph}, for dealing with both global (Section \ref{sec:globalgraphenc}) and local (Section \ref{sec:localgraphenc}) node contexts. We adopt GAT because it is closely related to the Transformer architecture \cite{NIPS2017_7181}, which provides a convenient prototype for modeling global node context. Then, (iii) we proposed strategies to combined the global and local graph encoders (Section
~\ref{sec:combiningenc}). Finally, (iv) we describe the decoding and training procedures (Section~\ref{sec:lp}).







\subsection{Graph Preparation}
\label{sec:graph_preparation}
We represent a KG as a multi-relational graph\footnote{In this paper, multi-relational graphs refer to directed graphs with labelled edges.} $\mathcal{G}_e = (\mathcal{V}_e, \mathcal{E}_e, \mathcal{R})$ with entity nodes $e \in \mathcal{V}_e$ and labeled edges $ (e_h, r, e_t) \in \mathcal{E}_e$, where $r \in \mathcal{R}$ denotes the relation existing from the entity $e_h$ to $e_t$.\footnote{$\mathcal{R}$ contains relations both in canonical direction (e.g. used-for) and in inverse direction (e.g. used-for-inv), so that the models consider the differences in the incoming and outgoing relations.}

Unlike other current approaches \cite{koncel-kedziorski-etal-2019-text, moryossef-etal-2019-step}, we represent an entity as a set of nodes. For instance, the KG node "node embedding" in Figure~\ref{fig:kg} will be represented by two nodes, one for the token "node" and the other for the token "embedding". Formally, we transform each $\mathcal{G}_e$ into a new graph $\mathcal{G} = (\mathcal{V}, \mathcal{E}, \mathcal{R})$, where each token of an entity $e \in \mathcal{V}_e$ becomes a node $v \in \mathcal{V}$. We convert each edge $ (e_h, r, e_t) \in \mathcal{E}_e$ into a set of edges (with the same relation $r$) and connect every token of $e_h$ to every token of $e_t$. That is, an edge $(u, r, v)$ will belong to $\mathcal{E}$ if and only if there exists an edge $ (e_h, r, e_t) \in \mathcal{E}_e$ such that $u \in e_h$ and $v \in e_t$, where $e_h$ and $e_t$ are seen as sets of tokens. We represent each node $v \in \mathcal{V}$ with an embedding $h^{0}_v \in  \mathbb{R}^{d_v}$, generated from its corresponding token.

The new graph $\mathcal{G}$ increases the representational power of the models because it allows learning node embeddings at a token level, instead of entity level. This is particularly important for text generation as it permits the model to be more flexible, capturing richer relationships between entity tokens. This also allows the model to learn relations and attention functions between source and target tokens. However, it has the side effect of removing the natural sequential order of multi-word entities. To preserve this information, we employ position embeddings \cite{NIPS2017_7181}, i.e., $h^{0}_v$ becomes the sum of the corresponding token embedding and the positional embedding for $v$.

\subsection{Graph Neural Networks (GNN)}
Multi-layer GNNs work by iteratively learning a representation vector $h_v$ of a node $v$ based on both its context node neighbors and edge features, through an information propagation scheme. More formally, the $l$-th layer aggregates the representations of $v$'s context nodes:
$$
        h_{\mathcal{N}(v)}^{(l)} = \textrm{\small AGGR}^{(l)} \big( \big\{ \big( h_u^{(l-1)}, r_{uv} \big) : u \in \mathcal{N}(v)  \big\} \big)  \, ,
$$
where $\textrm{AGGR}^{(l)}(.)$ is an aggregation function, shared by all nodes on the $l$-th layer. $r_{uv}$ represents the relation between $u$ and $v$. $\mathcal{N}(v)$ is a set of context nodes for $v$. In most GNNs, the context nodes are those adjacent to $v$. $h_{\mathcal{N}(v)}^{(l)}$ is the aggregated context representation of $\mathcal{N}(v)$ at layer $l$. $h_{\mathcal{N}(v)}^{(l)}$ is used to update the representation of $v$:
$$
       h_{v}^{(l)} = \textrm{\small COMBINE}^{(l)} \Big( h_v^{(l-1)} \, , h_{\mathcal{N}(v)}^{(l)}  \Big) \, .
$$

After $L$ iterations, a node's representation encodes the structural information within its $L$-hop neighborhood.
The choices of $\textrm{AGGR}^{(l)}(.)$ and $\textrm{COMBINE}^{(l)}(.)$ differ by the specific GNN model. An example of $\textrm{AGGR}^{(l)}(.)$ is the sum of the representations of $\mathcal{N}(v)$. An example of $\textrm{COMBINE}^{(l)}(.)$ is a concatenation after the feature transformation. 




\subsection{Global Graph Encoder} 
\label{sec:globalgraphenc}

A global graph encoder aggregates the \textit{global context} for updating each node based on all nodes of the graph (see Figure \ref{fig:kg}c). We use the attention mechanism as the message passing scheme, extending the self-attention network structure of Transformer to a GAT structure. In particular, we compute a layer of the \textit{global convolution} for a node $v \in \mathcal{V}$, which takes the input feature representations $h_v$ as input, adopting $\textrm{AGGR}^{(l)}(.)$ as:

\begin{equation}
h_{\mathcal{N}(v)} = \sum\nolimits_{\, u \in \mathcal{V}} \, \alpha_{vu} \, W_{g} \, h_u \, ,
\label{eq:global1}
\end{equation}
where $W_{g} \in \mathbb{R}^{d_v \times d_z}$ is a model parameter. The attention weight $\alpha_{vu}$ is calculated as:
\begin{equation}
\alpha_{vu} = \frac{\exp(e_{vu})}{ \sum_{k \in \mathcal{V}} \exp(e_{vk})}\,\, ,
\end{equation}

\noindent where, 
\begin{equation}
e_{vu} = \Big( \big( W_{q} h_v \big)^{\top}  \big( W_{k} h_u \big) \Big) / d_z
\label{eq:global3}
\end{equation}
is the attention function which measures the \emph{global importance} of node $u$'s features to node $v$. $W_{q}, W_{k} \in \mathbb{R}^{d_v \times d_z}$ are model parameters and $d_z$ is a scaling factor. 
To capture distinct relations between nodes, $K$ independent global convolutions are calculated and concatenated:
\begin{equation}
\hat{h}_{\mathcal{N}(v)} = \bigparallel\nolimits_{\, k=1}^{\, K} h^{(k)}_{\mathcal{N}(v)} \,\, .
\label{eq:concat}
\end{equation}


 \begin{figure*}[t]
    \centering
    \includegraphics[width=1\textwidth]{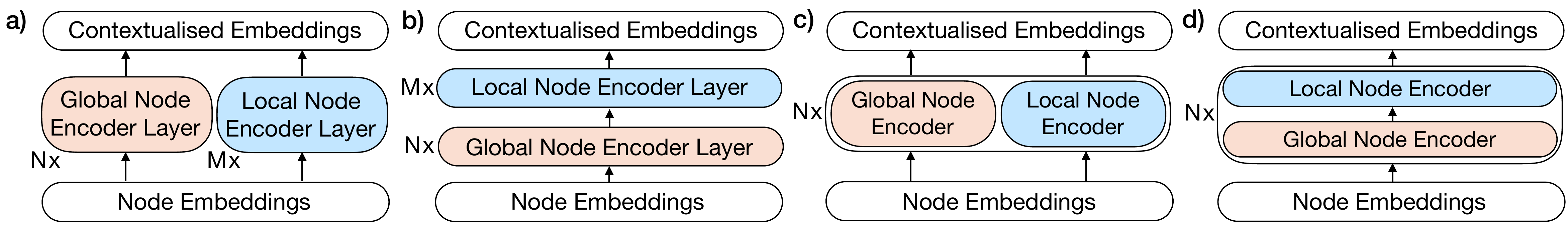}
    \vspace{-7mm}
    \caption{Overview of the proposed encoder architectures. (a) Parallel Graph Encoder ({\fontfamily{qcr}\selectfont PGE}) with separated parallel global and local node encoders. (b) Cascaded Graph Encoder ({\fontfamily{qcr}\selectfont CGE}) with separated cascaded encoders. c) {\fontfamily{qcr}\selectfont PGE-LW}: global and local node representations are concatenated layer-wise. d) {\fontfamily{qcr}\selectfont CGE-LW}: Both node representations are cascaded layer-wise.}
    \label{fig:encoder}
\end{figure*}

Finally, we define $\textrm{COMBINE}^{(l)}(.)$ employing layer normalization (LayerNorm) and a fully connected feed-forward network (FFN), in a similar way as the transformer architecture:
\begin{align}
  \hat{h}_{v} &= \text{LayerNorm}(\hat{h}_{\mathcal{N}(v)} + h_{v}) \, , \label{eq:2}\\
h^{global}_{v} &= \text{FFN}(\hat{h}_{v}) \, + \, \hat{h}_{\mathcal{N}(v)} + h_{v} \, \label{eq:2b}.
\end{align}



Note that the global encoder creates an artificial complete graph with $\mathcal{O}(n^2)$ edges and does not consider the edge relations. In particular, if the labelled edges were considered, the self-attention space complexity would increase to $ \Theta( |\mathcal{R}| \,  n^2)$.

\subsection{Local Graph Encoder} 
\label{sec:localgraphenc}
The representation $h^{global}_{v}$ captures macro relationships from $v$ to all other nodes in the graph. However, this representation lacks both structural information regarding the local neighborhood of $v$ and the graph topology. Also, it does not capture labelled edges (relations) between nodes (see Equations~\ref{eq:global1} and \ref{eq:global3}). 
In order to capture these crucial graph properties and impose a strong relational inductive bias, we build a graph encoder to aggregate the \emph{local context} by employing a modified version of GAT augmented with relational weights. 
In particular, we compute a layer of the \textit{local convolution} for a node $v \in \mathcal{V}$, adopting $\textrm{AGGR}^{(l)}(.)$ as:
\begin{equation}
        h_{\mathcal{N}(v)} =
        \sum\nolimits_{\, u \, \in \, \mathcal{N}(v)} \alpha_{vu} \, W_r \, h_u \, ,
\label{eq:3}
\end{equation}
where $W_{r} \in \mathbb{R}^{d_v \times d_z}$ encodes the relation $r \in \mathcal{R}$ between $u$ and $v$. $\mathcal{N}(v)$ is a set of nodes adjacent to $v$ and $v$ itself. The attention coefficient $\alpha_{vu}$ is computed as:
\begin{equation}
\alpha_{vu} = \frac{\exp(e_{vu})}{ \sum_{k \, \in \, \mathcal{N}(v)} \exp(e_{vk})} \, ,
\label{eq:4}
\end{equation}
\noindent where, 
\begin{equation}
e_{vu} = \sigma \big( a^{\top}
        [ W_v h_v \, \Vert \, W_r h_u]
        \big)
\label{eq:5}
\end{equation}
is the attention function which calculates the \emph{local importance} of adjacent nodes, considering the edge labels. $\sigma$ is an activation function, $\Vert$ denotes concatenation and $W_{v} \in \mathbb{R}^{d_v \times d_z}$ and $a \in \mathbb{R}^{2d_z}$ are model parameters. 

We employ multi-head attentions to learn local relations in different perspectives, as in Equation~\ref{eq:concat}, generating $\hat{h}_{\mathcal{N}(v)}$.
Finally, we define $\textrm{COMBINE}^{(l)}(.)$ as:
\begin{equation}
h_v^{local} = \textrm{\small RNN}(h_v, \hat{h}_{\mathcal{N}(v)}) \, ,
\end{equation}
where we employ as RNN a Gated Recurrent Unit (GRU) \cite{cho-etal-2014-learning}. GRU facilitates information propagation between local layers. This choice is motivated by recent works \cite{Xu2018RepresentationLO, NIPS2019_9675} that theoretically demonstrate that sharing information between layers helps the structural signals propagate. In a similar direction, AMR-to-text generation models employ LSTMs \cite{song-etal-2017-amr} and dense connections \cite{dcgcnforgraph2seq19guo} between GNN layers.



\begin{table*}[t]
\centering
\small
{\renewcommand{\arraystretch}{0.8}

\begin{tabular}{@{\hspace*{2mm}}c@{\hspace*{2mm}}r@{\hspace*{2mm}}r@{\hspace*{2.5mm}}r@{\hspace*{2.5mm}}r@{\hspace*{2.5mm}}r@{\hspace*{2.5mm}}r@{\hspace*{2.5mm}}r@{\hspace*{2.5mm}}r@{\hspace*{2.5mm}}r}  
\toprule
 & \textbf{\#train} & \textbf{\#dev} & \textbf{\#test} & \textbf{\#relations} & \textbf{avg \#entities} & \textbf{avg \#nodes} & \textbf{avg \#edges}  & \textbf{avg \#CC} & \textbf{avg length}  \\
\midrule
AGENDA & 38,720 & 1,000 & 1,000 & 7\;\;\;\; & 12.4\;\;\;\;\;\; & 44.3\;\;\;\;\; & 68.6\;\;\;\;\; & 19.1\;\;\;\; & 140.3\;\;\;\;  \\
WebNLG & 18,102 & 872 & 971 & 373\;\;\;\; & 4.0\;\;\;\;\;\; & 34.9\;\;\;\;\; & 101.0\;\;\;\;\; & 1.5\;\;\;\; & 24.2\;\;\;\;  \\
\bottomrule
\end{tabular}}
\vspace{-3.5mm}
\caption{Data statistics. Nodes, edges and CC values are calculated after the graph transformation. The average values are calculated for all splits (training, dev and test sets). CC refers to the number of connected components.}
\vspace{-3mm}
\label{tab:datastatistics}
\end{table*}


\subsection{Combining Global and Local Encodings}
\label{sec:combiningenc}

Our goal is to implement a graph encoder capable of encoding global and local aspects of the input graph. We hypothesize that these two sources of information are complementary, and a combination of both enriches node representations for text generation. In order to test this hypothesis, we investigate different combined architectures. 


Intuitively, there are two general methods for integrating two types of representation. The first is to concatenate vectors of global and local contexts, which we call a {\it parallel} representation. The second is to form a pipeline, where a global representation is first obtained, which is then used as a input for calculating refined representations based on the local node context. We call this approach a {\it cascaded} representation. 

Parallel and cascaded integration can be performed at the model level, considering the global and local graph encoders as two representation learning units disregarding internal structures. However, because our model takes a multi-layer architecture, where each layer makes a level of abstraction in representation, we can alternatively consider integration on the layer level, so that more interaction between global and local contexts may be captured. As a result, we present four architectures for integration, as shown in Figure~\ref{fig:encoder}. All models serve the same purpose, and their relative strengths should be evaluated empirically.

\paragraph{Parallel Graph Encoding (PGE).} In this setup, we compose global and local graph encoders in a fully parallel structure (Figure~\ref{fig:encoder}a). Note that each graph encoder can have different numbers of layers and attention heads. The final node representation is the concatenation of the local and global node representations of the last layers of both graph encoders:
\begin{align}
h^{global}_{v} &= \textrm{GE} (h^{0}_v, \{ h^{0}_u : u \in \mathcal{V} \}) \, \nonumber \\
h^{local}_{v} &= \textrm{LE} (h^{0}_v, \{ h^{0}_u : u \in \mathcal{N}(v) \}) \, \nonumber \\
h_{v} &= [ \, h^{global}_{v} \, \Vert \, h^{local}_{v} \, ] \, ,
\end{align}
where $\textrm{GE}$ and $\textrm{LE}$ denote the global and local graph encoders, respectively. $h^{0}_v$ is the initial node embedding used in the first layer of both encoders.
\paragraph{Cascaded Graph Encoding (CGE).} We cascade local and global graph encoders as shown in Figure~\ref{fig:encoder}b. We first compute a globally contextualized node embedding, and then refining it with the local node context. $h^{0}_v$ is the initial input for the global encoder and $h^{global}_{v}$ is the initial input for the local encoder. In particular, the final node representation is calculated as follows:

\begin{align}
h^{global}_{v} &= \textrm{GE} (h^{0}_v, \{ h^{0}_u : u \in \mathcal{V} \}) \, \nonumber \\
h_{v} &= \textrm{LE} (h^{global}_v,\!\{ h^{global}_u\!: u \in \mathcal{N}(v) \}).\hspace{-.5em}
\end{align}

\paragraph{Layer-wise Parallel Graph Encoding.} To allow fine-grained interaction between the two types of graph contextual information, we also combine the encoders in a layer-wise (LW) fashion. As shown in Figure~\ref{fig:encoder}c, for each graph layer, we employ both global and local encoders in a parallel structure ({\fontfamily{qcr}\selectfont PGE-LW}). More precisely, each encoder layer is calculates as follows: 
\begin{align}
h^{global}_{v} &= \textrm{GE}_{l} (h^{l - 1}_v, \{ h^{l - 1}_u : u \in \mathcal{V} \}) \, \nonumber \\
h^{local}_{v} &= \textrm{LE}_{l} (h^{l - 1}_v, \{ h^{l - 1}_u : u \in \mathcal{N}(v) \}) \, \nonumber \\
h^{l}_{v} &= [ \, h^{global}_{v} \, \Vert \, h^{local}_{v} \, ] \, ,
\end{align}
where $\textrm{GE}_{l}$ and $\textrm{LE}_{l}$ refer to the $l$-th layers of the global and local graph encoders, respectively.
\paragraph{Layer-wise Cascaded Graph Encoding.} We also propose cascading the graph encoders layer-wise ({\fontfamily{qcr}\selectfont CGE-LW}, Figure~\ref{fig:encoder}d). In particular, we compute each encoder layer as follows: 
\begin{fleqn}
\begin{align}
h^{global}_{v} &= \textrm{GE}_{l} (h^{l - 1}_v, \{ h^{l - 1}_u : u \in \mathcal{V} \}) \, \nonumber \\
h^{l}_{v} &= \textrm{LE}_{l} (h^{global}_v,\!\{ h^{global}_u\!: u \in \mathcal{N}(v) \}).\hspace{-.5em}
\end{align}
\end{fleqn}



\vspace{0.5mm}
\subsection{Decoder and Training}
\label{sec:lp}
Our decoder follows the core architecture of a Transformer decoder \cite{NIPS2017_7181}. Each time step $t$ is updated by performing multi-head attentions over the output of the encoder (node embeddings $h_v$) and over previously-generated tokens (token embeddings). An additional challenge in our setup is to generate multi-sentence outputs. In order to encourage the model to generate longer texts, we employ a length penalty \cite{DBLP:journals/corr/WuSCLNMKCGMKSJL16} to refine the pure max-probability beam search. 

The model is trained to optimize the negative log-likelihood of each gold-standard output text. We employ label smoothing regularization to prevent the model from predicting the tokens too confidently during training and generalizing poorly.




\section{Data and Preprocessing}

We attest the effectiveness of our models on two datasets: AGENDA \cite{koncel-kedziorski-etal-2019-text} and WebNLG \cite{gardent-etal-2017-webnlg}. Table ~\ref{tab:datastatistics} shows the statistics for both datasets.

\begin{table*}[h]
\centering
{\renewcommand{\arraystretch}{0.8}
\begin{tabular}{lcccccc}  
\toprule
\textbf{Model} & \textbf{\#L} & \textbf{\#H} & \textbf{BLEU} & \textbf{METEOR} & \textbf{CHRF++} & \textbf{\#P}   \\
\midrule
 \citet{koncel-kedziorski-etal-2019-text} & 6 & 8 & 14.30 {\small $\pm$1.01}& 18.80 {\small $\pm$0.28} & - & - \\
\midrule
 Global Encoder & 6 & 8 & 15.44 {\small $\pm$0.25} & 20.76 {\small $\pm$0.19} & 43.95 {\small $\pm$0.40} & 54.4 \\
 Local Encoder & 3 & 8 & 16.03 {\small $\pm$0.19} & 21.12 {\small $\pm$0.32} & 44.70 {\small $\pm$0.29} & 54.0 \\
  PGE & 6, 3 & 8, 8 & 17.55 {\small $\pm$0.15} & 22.02 {\small $\pm$0.07} & \textbf{46.41} {\small $\pm$0.07}& 56.1 \\
 CGE & 6, 3 & 8, 8 & \textbf{17.82} {\small $\pm$0.13} & \textbf{22.23} {\small $\pm$0.09} & \textbf{46.47} {\small $\pm$0.10} & 61.5 \\
   PGE-LW  & 6 & 8, 8 & 17.42 {\small $\pm$0.25} & 21.78 {\small $\pm$0.20} & 45.79 {\small $\pm$0.32} & 69.0 \\
  CGE-LW & 6 & 8, 8 & \textbf{18.01} {\small $\pm$0.14} & \textbf{22.34} {\small $\pm$0.07} & \textbf{46.69} {\small $\pm$0.17} & 69.8 \\
\bottomrule
\end{tabular}}
\caption{Results on AGENDA test set. \#L and \#H are the numbers of layers and the attention heads in each layer, respectively. When more than one, the values are for the global and local encoders, respectively. \#P stands for the number of parameters in millions (node embeddings included).}
\label{tab:AGENDAtestresults}
\end{table*}


\paragraph{AGENDA.} In this dataset, KGs are paired with scientific abstracts extracted from proceedings of 12 top AI conferences. Each instance consists of the paper title, a KG and the paper abstract. Entities correspond to scientific terms which are often multi-word expressions (co-referential entities are merged). We treat each token in the title as a node, creating a unique graph with title and KG tokens as nodes. As shown in Table~\ref{tab:datastatistics}, the average output length is considerably large, as the target outputs are multi-sentence abstracts. 
\paragraph{WebNLG.} In this dataset, each instance contains a KG extracted from DBPedia. The target text consists of sentences that verbalise the graph. We evaluate the models on the test set with seen categories. Note that this dataset has a considerable number of edge relations (see Table~\ref{tab:datastatistics}). In order to avoid parameter explosion, we use regularization based on the basis function decomposition to define the model relation weights \cite{Schlichtkrull2018ModelingRD}. Also, as an alternative, we employ the Levi Transformation to create nodes from relational edges between entities \cite{beck-etal-2018-acl2018}. That is, we create a new relation node for each edge relation between two nodes. The new relation node is connected to the subject and object token entities by two binary relations, respectively. 



\section{Experiments}

\begin{figure*}[t]
    \centering
    \includegraphics[width=1\textwidth]{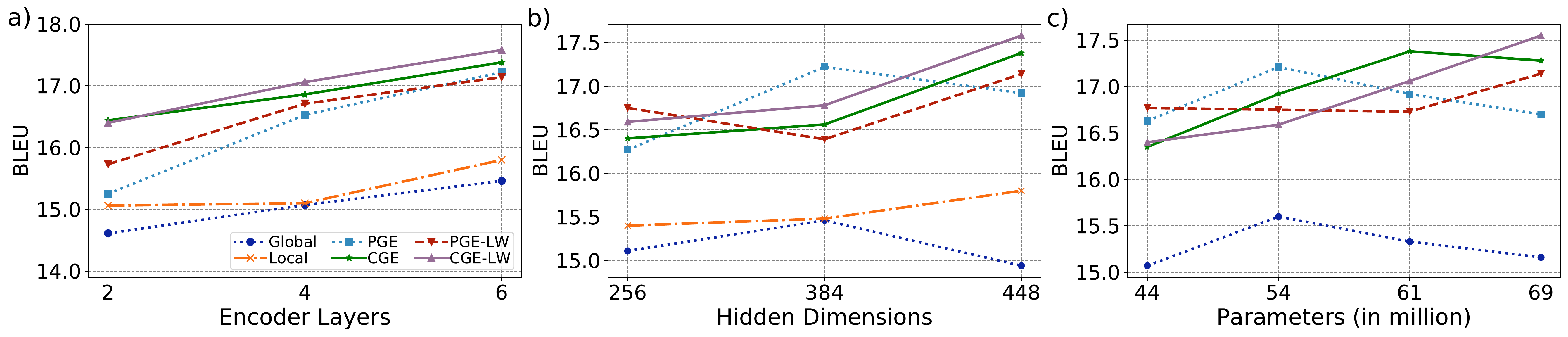}
    \vspace{-7mm}
    \caption{BLEU scores for AGENDA dev set, with respect to (a) the encoder layers, (b) the encoder hidden dimensions and (c) the number of parameters.}
    \label{fig:analysis_arcs}
\end{figure*}

We implemented all our models using PyTorch Geometric (PyG) \cite{Fey/Lenssen/2019} and \mbox{OpenNMT-py} \cite{opennmt}. We employ the Adam optimizer with $\beta_1=0.9$ and $\beta_2=0.98$. Our learning rate schedule follows \citet{NIPS2017_7181} with 8000 and 16000 warming-up steps for WebNLG and AGENDA, respectively. The vocabulary is shared between the node and target tokens. In order to mitigate the effects of random seeds, for the test sets, we report the averages over 4 training runs along with their standard deviation. We employ byte pair encoding (BPE, \citeauthor{sennrich-etal-2016-neural}, \citeyear{sennrich-etal-2016-neural}) to split entity words into smaller more frequent pieces. So some nodes in the graph can be sub-words. We also obtain sub-words on the target side. Following previous works, we evaluate the results with BLEU \cite{Papineni:2002:BMA:1073083.1073135}, METEOR \cite{Denkowski14meteoruniversal} and CHRF++ \cite{popovic-2015-chrf} automatic metrics and also perform a human evaluation (Section~\ref{sec:humaneval}). For layer-wise models, the number of encoder layers are chosen from $\{2, 4, 6\}$, and for {\fontfamily{qcr}\selectfont PGE} and {\fontfamily{qcr}\selectfont CGE}, the global and local layers are chosen from and $\{2, 4, 6\}$ and $\{1, 2, 3\}$, respectively. The hidden encoder dimensions are chosen from $\{256, 384, 448\}$ (see Figure~\ref{fig:analysis_arcs}). Hyperparameters are tuned on the development set of both datasets. We report the test results when the BLEU score on dev set is optimal.

\begin{table*}[t]
\centering
{\renewcommand{\arraystretch}{0.8}
\begin{tabular}{lllll}  
\toprule
\textbf{Model} & \textbf{BLEU} & \textbf{METEOR} & \textbf{CHRF++} &  \textbf{\#P}  \\
\midrule
 UPF-FORGe \cite{gardent-etal-2017-webnlg}   & 40.88 & 40.00 & -  & - \\
 Melbourne \cite{gardent-etal-2017-webnlg}  & 54.52 & 41.00 & 70.72 & -  \\
 Adapt \cite{gardent-etal-2017-webnlg}  & 60.59 & 44.00 & 76.01 & - \\
 \citet{marcheggiani-icnl18}   & 55.90 & 39.00 & - & 4.9 \\
  \citet{trisedya-etal-2018-gtr}   & 58.60 & 40.60 & - & - \\
 \citet{castro-ferreira-etal-2019-neural}  & 57.20 & 41.00 & -  & -\\
\midrule
 CGE & 62.30 {\small $\pm$0.27} & 43.51 {\small $\pm$0.18} & 75.49 {\small $\pm$0.34}  & 13.9 \\
 CGE (Levi Graph)  & 63.10 {\small $\pm$0.13} & 44.11 {\small $\pm$0.09} & 76.33 {\small $\pm$0.10}  & 12.8 \\
  CGE-LW & 62.85 {\small $\pm$0.07} & 43.75 {\small $\pm$0.21} & 75.73 {\small $\pm$0.31}  & 11.2 \\
 CGE-LW (Levi Graph)  & \textbf{63.69} {\small $\pm$0.10} & \textbf{44.47} {\small $\pm$0.12} & \textbf{76.66} {\small $\pm$0.10}  & 10.4 \\
\bottomrule
\end{tabular}}
\caption{Results on WebNLG test set with seen categories.}
\label{tab:webnlg_results}
\end{table*}

\subsection{Results on AGENDA}
Table~\ref{tab:AGENDAtestresults} shows the results, where we report the number of layers and attention heads employed. We train models with only global or local encoders as baselines. Each model has the respective parameter size that gives the best results on the dev set. First, the local encoder, which requires fewer encoder layers and parameters, has a better performance compared to the global encoder. This shows that explicitly encoding the graph structure is important to improve the node representations. Second, our approaches substantially outperform both baselines. {\fontfamily{qcr}\selectfont CGE-LW} outperforms \citet{koncel-kedziorski-etal-2019-text}, a transformer model that focuses on the relations between adjacent nodes, by a large margin, achieving the new state-of-the-art BLEU score of 18.01, 25.9\% higher. We also note that KGs are highly incomplete in this dataset, with an average number of connected components of 19.1 (see Table~\ref{tab:datastatistics}). For this reason, the global encoder plays an important role in our models as it enables learning node representations based on all connected components. The results indicate that combining the local node context, leveraging the graph topology, and the global node context, capturing macro-level node relations, leads to better performance. We find that, even though {\fontfamily{qcr}\selectfont CGE} has a small number of parameters compared to {\fontfamily{qcr}\selectfont CGE-LW}, it achieves comparable performance. {\fontfamily{qcr}\selectfont PGE-LW} has the worse performance among the proposed models. Finally note that cascaded architectures are more effective according to different metrics.

 \begin{figure*}[t]
    \centering
    \includegraphics[width=1\textwidth]{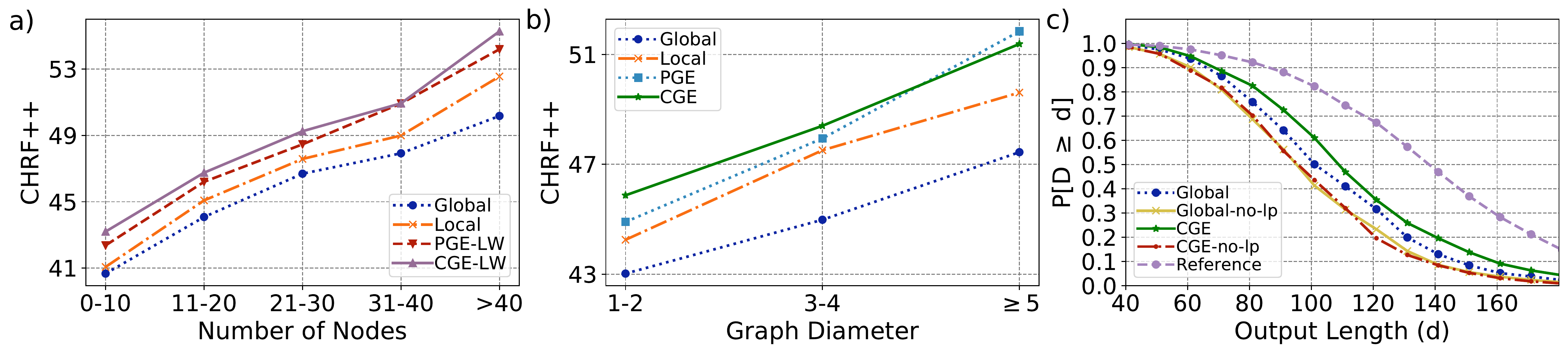}
    \vspace{-7mm}
    \caption{CHRF++ scores for AGENDA test set, with respect to (a) the number of nodes, and (b) the graph diameter. (c) Distribution of length of the gold references and models' outputs for the AGENDA test set.}
    \label{fig:analysis1}
\end{figure*}


\subsection{Results on WebNLG}

We compare the performance of our more effective models ({\fontfamily{qcr}\selectfont CGE}, {\fontfamily{qcr}\selectfont CGE-LW}) with six state-of-the-art results reported on this dataset. Three systems are the best competitors in the WebNLG challenge for seen categories: UPF-FORGe, Melbourne and Adapt. UPF-FORGe follows a rule-based approach, whereas the others use neural encoder-decoder models with linearized triple sets as input. 

Table~\ref{tab:webnlg_results} presents the results. {\fontfamily{qcr}\selectfont CGE} achieves a BLEU score of 62.30, 8.9\% better than the best model of \citet{castro-ferreira-etal-2019-neural}, who employ an end-to-end architecture based on GRUs. {\fontfamily{qcr}\selectfont CGE} using Levi graphs outperforms \citet{trisedya-etal-2018-gtr}, an approach that encodes both intra-triple and inter-triple relationships, by 4.5 BLEU points. Interestingly, their intra-triple and inter-triple mechanisms are closely related with the local and global encodings. However, they rely on encoding entities based on sequences generated by traversal graph algorithms, whereas we explicitly exploit the graph structure, throughout the local neighborhood aggregation. 

{\fontfamily{qcr}\selectfont CGE-LW} with Levi graphs as inputs has the best performance, achieving 63.69 BLEU points, even thought it uses fewer parameters. Note that this approach allows the model to handle new relations, as they are treated as nodes. Moreover, the relations become part of the shared vocabulary, making this information directly usable during the decoding phase. We outperform an approach based on GNNs \cite{marcheggiani-icnl18} by a large margin of 7.7 BLEU points, showing that our combined graph encoding strategies lead to better text generation. We also outperform Adapt, a strong competitor that employs subword encodings, by 3.1 BLEU points.


\subsection{Development Experiments}

We report several development experiments in Figure~\ref{fig:analysis_arcs}. Figure~\ref{fig:analysis_arcs}a shows the effect of the number of encoder layers in the four encoding methods.\footnote{For {\fontfamily{qcr}\selectfont CGE} and {\fontfamily{qcr}\selectfont PGE} the values refer to the global layers and the number of local layers is fixed to 3.}  In general, the performance increases when we gradually enlarge the number of layers, achieving the best performance with 6 encoder layers. Figure~\ref{fig:analysis_arcs}b shows the choices of hidden sizes for the encoders. The best performances for global and {\fontfamily{qcr}\selectfont PGE} are achieved with 384 dimensions, whereas the other models have the better performance with 448 dimensions. In Figure~\ref{fig:analysis_arcs}c, we evaluate the performance employing different number of parameters.\footnote{It was not possible to execute the local model with larger number of parameters due to memory limitations.} When the models are smaller, parallel encoders obtain better results than the cascaded ones. When the models are larger, cascaded models perform better. We speculate that for some models, the performance can be further improved with more parameters and layers. However, we do not attempt this owing to hardware limitations.


\subsection{Ablation Study} In Table~\ref{tab:ablation}, we report an ablation study on the impact of each module used in {\fontfamily{qcr}\selectfont CGE} model on the dev set of AGENDA. We also report the number of parameters used in each configuration. 
\paragraph{Global Graph Encoder.} We start by an ablation on the global encoder. After removing the global attention coefficients, the performance of the model drops by 1.79 BLEU and 1.97 CHRF++ scores. Results also show that using FFN in the global $\textrm{COMBINE}(.)$ function is important to the model but less effective than the global attention. However, when we remove FNN, the number of parameters drops considerably (around 18\%) from 61.5 to 50.4 million. Finally, without the entire global encoder, the result drops substantially by 2.21 BLEU points. This indicates that enriching node embeddings with a global context allows learning more expressive graph representations.

\begin{table}[t]
\centering
{\renewcommand{\arraystretch}{0.8}
\setlength{\belowrulesep}{1.3pt}
\setlength{\aboverulesep}{0pt}
\begin{tabular}{@{\hspace*{1.8mm}}lccc@{\hspace*{1.8mm}}}  
\toprule
\textbf{Model} & \textbf{BLEU} & \textbf{CHRF++} &  \textbf{\#P}   \\
\midrule
CGE & 17.38 & 45.68 & 61.5 \\
\midrule
Global Encoder &  & &  \\
 -Global Attention & 15.59 & 43.71 & 59.0 \\
 -FFN & 16.33 & 44.86 & 50.4 \\
 -Global Encoder & 15.17 & 43.30 & 45.6 \\
\midrule
Local Encoder &  & &  \\
 -Local Attention & 16.92 & 45.97 & 61.5 \\
 -Weight Relations & 16.88 & 45.61 & 53.6 \\
 -GRU & 16.38 & 44.71 & 60.2 \\
 -Local Encoder & 14.68 & 42.98 & 51.8  \\
 \midrule
 -Shared Vocab. & 16.92 & 46.16 & 81.8 \\
 \midrule
 Decoder &  & & \\
 -- Length Penalty & 16.68 & 44.68 & 61.5 \\
\bottomrule
\end{tabular}}
\caption{Ablation study for modules used in the encoder and decoder of the {\fontfamily{qcr}\selectfont CGE} model.}
\label{tab:ablation}
\end{table}


\paragraph{Local Graph Encoder.} We first remove the local graph attention and the BLEU score drops to 16.92, showing that the neighborhood attention improves the performance. After removing the relation types, encoded as model weights, the performance drops by 0.5 BLEU points. However, the number of parameters is reduced by around 7.9 million. This indicates that we can have a more efficient model, in terms of the number of parameters, with a slight drop in performance. Removing the GRU used on the $\textrm{COMBINE}(.)$ function decreases the performance considerably. The worse performance occurs if we remove the entire local encoder, with a BLEU score of 14.68, essentially making the encoder similar to the global baseline.

Finally, we find that vocabulary sharing improves the performance, and the length penalty is beneficial as we generate multi-sentence outputs.

\subsection{Impact of the Graph Structure and Output Length}
\label{sec:outputlength}
The overall performance on both datasets suggests the strength of combining global and local node representations. However, we are also interested in estimating the models' performance concerning different data properties.
\paragraph{Graph Size.} Figure~\ref{fig:analysis1}a shows the effect of the graph size, measured in number of nodes, on the performance, measured using CHRF++ scores,\footnote{CHRF++ score is used as it is a sentence-level metric.} for the AGENDA. We evaluate global and local graph encoders, {\fontfamily{qcr}\selectfont PGE-LW} and {\fontfamily{qcr}\selectfont CGE-LW}. We find that the score increases as the graph size increases. Interesting, the gap between the local and global encoders increases when the graph size increases. This suggests that, because larger graphs may have very different topologies, modeling the relations between nodes based on the graph structure is more beneficial than allowing direct communication between nodes, overlooking the graph structure. Also note that the the cascaded model ({\fontfamily{qcr}\selectfont CGE-LW}) is consistently better than the parallel model ({\fontfamily{qcr}\selectfont PGE-LW}) over all graph sizes.

\begin{table}
\centering
{\renewcommand{\arraystretch}{0.8}
\setlength{\belowrulesep}{1.3pt}
\setlength{\aboverulesep}{1pt}
\begin{tabular}{@{\hspace*{1.5mm}}c@{\hspace*{2mm}}r@{\hspace*{2mm}}c@{\hspace*{5mm}}c@{\hspace*{5mm}}c@{\hspace*{1.5mm}}}  
\toprule
\textbf{\#T} & \textbf{\#DP} & \textbf{Melbourne} & \textbf{Adapt} & \textbf{CGE-LW}   \\
\midrule
1-2 & 396 & 78.74 & 83.10 & 84.35 \\
3-4 & 386 & 66.84 & 72.02 & 72.27 \\
5-7 & 189 & 61.85 & 69.28 & 70.25 \\
\midrule
\textbf{\#D} & \textbf{\#DP} & \textbf{Melbourne} & \textbf{Adapt} & \textbf{CGE-LW}   \\
\midrule
1 & 222 & 82.27 & 87.54 & 88.04 \\
2 & 469 & 69.94 & 74.54 & 75.90 \\
$\geq3$ & 280 & 62.87 & 69.30 & 69.41 \\
\midrule
\textbf{\#S} & \textbf{\#DP} & \textbf{Melbourne} & \textbf{Adapt} & \textbf{CGE-LW}   \\
\midrule
1 & 388 & 77.19 & 81.66 & 82.03  \\
2 & 306 & 67.29 & 73.29 & 73.78 \\
3 & 151 & 66.30 & 72.46 & 73.21 \\
4 & 66 & 66.73 & 71.26 & 75.16 \\
$\geq5$ & 60 & 61.93 & 67.57 & 69.20 \\
\bottomrule
\end{tabular}}
\caption{CHRF++ scores with respect to the number of triples (\#T), graph diameters (\#D) and number of sentences (\#S) on the WebNLG test set. \#DP refers to the number of datapoints.}
\label{tab:webnlg-stats}
\end{table}

Table~\ref{tab:webnlg-stats} shows the effect of the graph size, measured in number of triples, on the performance for the WebNLG. Our model obtains better scores over all partitions. In contrast to AGENDA, the performance decreases as the graph size increases. This behavior highlights a crucial difference between AGENDA and AMR and WebNLG datasets, in which the models' general performance decreases as the graph size increases \cite{gardent-etal-2017-webnlg,cai-lam-2020-graph}. In WebNLG, the graph and sentence sizes are correlated, and longer sentences are more challenging to generate than the smaller ones. Differently, AGENDA contains similar text lengths\footnote{As shown on Figure~\ref{fig:analysis1}c, 82\% of the reference abstracts have more than 100 words.} and when the input is a larger graph, the model has more information to be leveraged during the generation. 

\paragraph{Graph Diameter.} Figure~\ref{fig:analysis1}b shows the impact of the graph diameter\footnote{The diameter of a graph is defined as the length of the longest shortest path between two nodes. We convert the graphs into undirected graphs to calculate the diameters.} on the performance for the AGENDA. Similarly to the graph size, the score increases as the diameter increases. As the global encoder is not aware of the graph structure, this module has the worst scores, even though it enables direct node communication over long distance. In contrast, the local encoder can propagate precise node information throughout the graph structure for $k$-hop distances, making the relative performance better. Table~\ref{tab:webnlg-stats} shows the models' performances with respect to the graph diameter for WebNLG. Similarly to the graph size, the score decreases as the diameter increases.

 \begin{figure}[t]
    \centering
    \includegraphics[width=.43\textwidth]{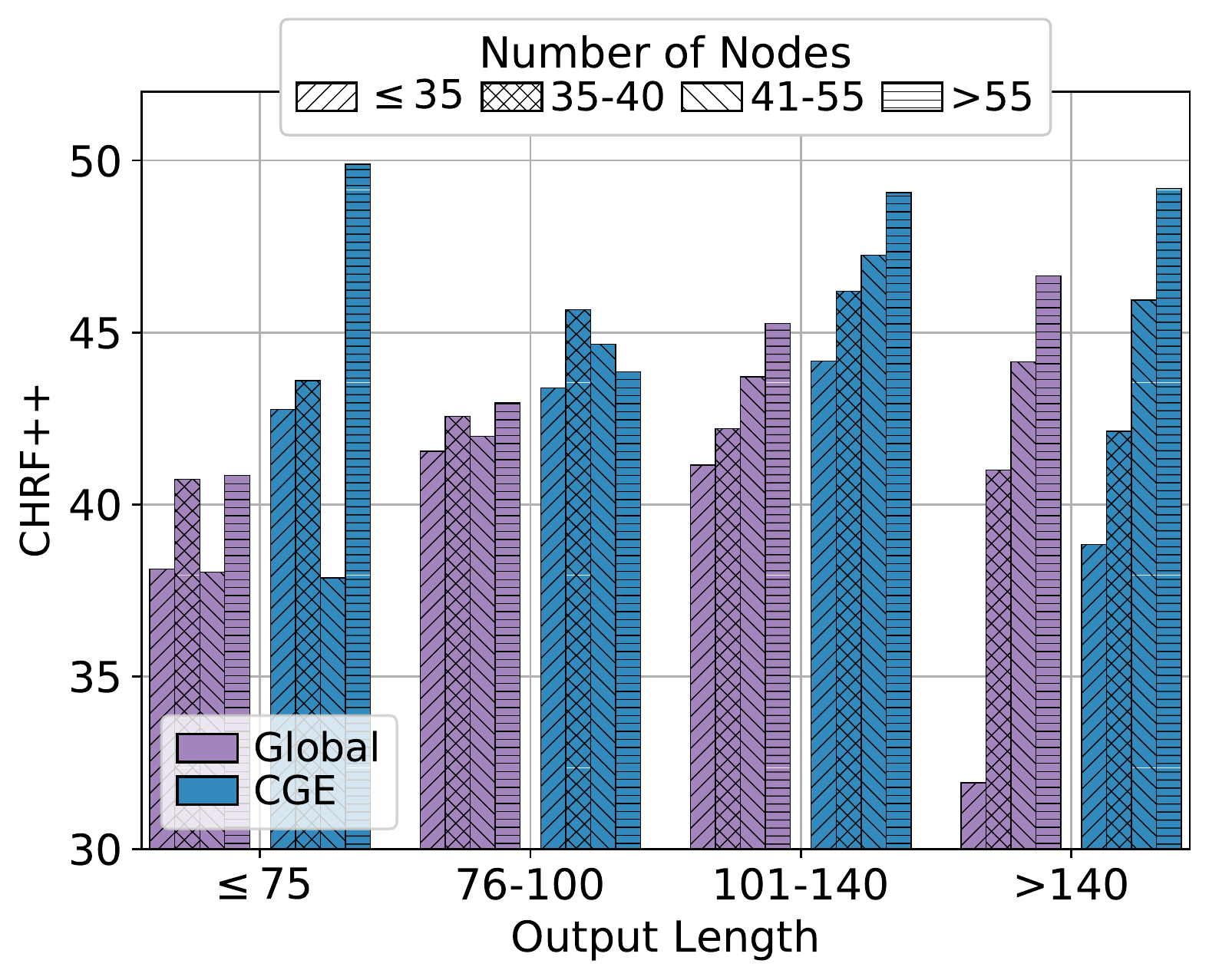}
    \vspace{-3mm}
    \caption{Relation between the number of nodes and the length of the generated text, in number of words.}
    \label{fig:comparison-input-output}
\end{figure}

\paragraph{Output Length.} One interesting phenomenon to analyze is the length distribution (in number of words) of the generated outputs. We expect that our models generate texts with similar output lengths as the reference texts. As shown in Figure~\ref{fig:analysis1}c, the references usually are bigger than the texts generated by all models for AGENDA. The texts generated by {\fontfamily{qcr}\selectfont CGE-no-pl}, a {\fontfamily{qcr}\selectfont CGE} model without length penalty, are consistently shorter than the texts from the global and {\fontfamily{qcr}\selectfont CGE} models. We increase the length of the texts when we employ the length penalty (see Section~\ref{sec:lp}). However, there is still a gap between the reference and the generated text lengths. We leave further investigation of this aspect for future work. 

Table~\ref{tab:webnlg-stats} shows the models' performances with respect to the number of sentences for WebNLG. In general, increasing the number of sentences reduces the performance of all models. Note that when the number of sentences increases, the gap between {\fontfamily{qcr}\selectfont CGE-LW} and the baselines becomes larger. This suggests that our approach is able to better handle complex graph inputs in order to generate multi-sentence texts.


\paragraph{Effect of the Number of Nodes on the Output Length.}
\label{sec:relations_input_output}
Figure~\ref{fig:comparison-input-output} shows the effect of the size of a graph, defined as the number of nodes, on the quality (measured in CHRF++ scores) and length of the generated text (in number of words) in the AGENDA dev set. We bin both the graph size and the output length in 4 classes. {\fontfamily{qcr}\selectfont CGE} consistently outperforms the global model, in some cases by a large margin. When handling smaller graphs (with $\leq35$ nodes), both models have difficulties generating good summaries. However, for these smaller graphs, our model achieves a score 12.2\% better when generating texts with length $\leq75$. Interestingly, when generating longer texts (>140) from smaller graphs, our model outperforms the global encoder by an impressive 21.7\%, indicating that our model is more effective in capturing semantic signals from graphs with scarce information. Our approach also performs better when the graph size is large ($> 55$) but the generation output is small ($\leq75$), beating the global encoder by 9 points.

\begin{table}[t]
\centering
{\renewcommand{\arraystretch}{0.8}
\setlength{\belowrulesep}{1.3pt}
\setlength{\aboverulesep}{0pt}
\begin{tabular}{@{\hspace*{5mm}}lcc@{\hspace*{5mm}}}  
\toprule
\multicolumn{3}{c}{\textbf{AGENDA}}  \\
\midrule
\textbf{Model} & \textbf{BLEU} & \textbf{CHRF++}  \\
\midrule
CGE-LW & 18.17 & 46.80 \\
 -Shared Vocab & 17.88 & 47.12 \\
 -Length Penalty & 17.46 & 45.76 \\
 -Both & 17.24 & 46.14 \\
 \midrule
\multicolumn{3}{c}{\textbf{WebNLG}}  \\
\midrule
\textbf{Model} & \textbf{BLEU} & \textbf{CHRF++}  \\
\midrule
CGE-LW & 63.86 & 76.80 \\
 -Shared Vocab & 63.07 & 76.17 \\
 -Length Penalty & 63.28 & 76.51 \\
 -Both & 62.60 & 75.80 \\
\bottomrule
\end{tabular}}
\vspace{-1.7mm}
\caption{Effects of the vocabulary sharing and length penalty on the test sets of AGENDA and WebNLG.}
\label{tab:ablation-test}
\end{table}

\subsection{Human Evaluation}
\label{sec:humaneval}
To further assess the quality of the generated text, we conduct a human evaluation on the WebNLG dataset.\footnote{Because AGENDA is scientific in nature, we choose to crowd source human evaluations only for WebNLG.} Following previous works \cite{gardent-etal-2017-webnlg, castro-ferreira-etal-2019-neural}, we assess two quality criteria: (i) \emph{Fluency} (i.e., does the text flow in a natural, easy to read manner?) and (ii) \emph{Adequacy} (i.e., does the text clearly express the data?). We divide the datapoints into seven different sets by the number of triples. For each set, we randomly select 20 texts generated by Adapt, {\fontfamily{qcr}\selectfont CGE} with Levi graphs and their corresponding human reference (420 texts in total). Since the number of datapoints for each set is not balanced (see Table~\ref{tab:webnlg-stats}), this sampling strategy assures us to have the same amount of samples for the different triple sets. Moreover, having human references may serve as an indicator of the sanity of the human evaluation experiment. We recruited human workers from Amazon Mechanical Turk to rate the text outputs on a 1-5 Likert scale. For each text, we collect scores from 4 workers and average them. Table~\ref{tab:humanevevaluation} shows the results. We first note a similar trend as in the automatic evaluation, with {\fontfamily{qcr}\selectfont CGE} outperforming Adapt on both fluency and adequacy. In sets with the number of triples smaller than 5, {\fontfamily{qcr}\selectfont CGE} was the highest rated system in fluency. Similarly to the automatic evaluation, both systems are better in generating text from graphs with smaller diameters. Note that bigger diameters pose difficulties to the models, which achieve their worst performance for diameters $\geq3$.

\begin{table}[t]
\centering
{\renewcommand{\arraystretch}{0.9}
\setlength\tabcolsep{2pt}
\setlength{\belowrulesep}{0pt}
\setlength{\aboverulesep}{0pt}
\begin{tabular}{@{\hspace*{0.2mm}}>{\centering}m{0.9cm} N{9mm} P{10mm} N{10mm} P{10mm} N{9mm} P{8mm} c} 
\toprule
\textbf{\#T} & \multicolumn{2}{c}{\textbf{Adapt}} & \multicolumn{2}{c}{\textbf{CGE}} & \multicolumn{2}{c}{\textbf{Reference}} &  \\
\midrule
\multicolumn{1}{@{\hspace*{0.2mm}}>{\centering\arraybackslash}m{0.9cm}}{} 
    & \multicolumn{1}{>{\centering\arraybackslash\columncolor{Gray}}m{9mm}}{F} 
    & \multicolumn{1}{>{\centering\arraybackslash}m{10mm}}{A}
    & \multicolumn{1}{>{\centering\arraybackslash\columncolor{Gray}}m{10mm}}{F} 
    & \multicolumn{1}{>{\centering\arraybackslash}m{10mm}}{A}
    & \multicolumn{1}{>{\centering\arraybackslash\columncolor{Gray}}m{9mm}}{F} 
    & \multicolumn{1}{>{\centering\arraybackslash}m{8mm}}{A}
    & \multicolumn{1}{>{\centering\arraybackslash}m{0mm}}{}\\
\midrule
 {\small All} &${3.96}^{{\scriptscriptstyle C}}$& ${4.44}^{{\scriptscriptstyle C}}$ & ${4.12}^{{\scriptscriptstyle B}}$ & ${4.54}^{{\scriptscriptstyle B}}$ & ${4.24}^{{\scriptscriptstyle A}}$ & ${4.63}^{{\scriptscriptstyle A}}$ &\\
\midrule
1-2 & ${3.94}^{{\scriptscriptstyle C}}$ & ${4.59}^{{\scriptscriptstyle B}}$ & ${4.18}^{{\scriptscriptstyle B}}$ & ${4.72}^{{\scriptscriptstyle A}}$ & ${4.30}^{{\scriptscriptstyle A}}$ & ${4.69}^{{\scriptscriptstyle A}}$ &\\
3-4 & ${3.79}^{{\scriptscriptstyle C}}$ & ${4.45}^{{\scriptscriptstyle B}}$ & ${3.96}^{{\scriptscriptstyle B}}$ & ${4.50}^{{\scriptscriptstyle AB}}$ & ${4.14}^{{\scriptscriptstyle A}}$ & ${4.66}^{{\scriptscriptstyle A}}$ &\\
5-7 & ${4.08}^{{\scriptscriptstyle B}}$ & ${4.35}^{{\scriptscriptstyle B}}$ & ${4.18}^{{\scriptscriptstyle B}}$ & ${4.45}^{{\scriptscriptstyle B}}$ & ${4.28}^{{\scriptscriptstyle A}}$ & ${4.59}^{{\scriptscriptstyle A}}$ &\\
\midrule
\textbf{\#D} & \multicolumn{2}{>{\centering}m{1.7cm}}{\textbf{Adapt}} & \multicolumn{2}{>{\centering}m{1.7cm}}{\textbf{CGE}} & \multicolumn{2}{>{\centering}m{1.7cm}}{\textbf{Reference}}  & \\
\midrule
\multicolumn{1}{@{\hspace*{0.2mm}}>{\centering\arraybackslash}m{0.9cm}}{} 
    & \multicolumn{1}{>{\centering\arraybackslash\columncolor{Gray}}m{9mm}}{F} 
    & \multicolumn{1}{>{\centering\arraybackslash}m{10mm}}{A}
    & \multicolumn{1}{>{\centering\arraybackslash\columncolor{Gray}}m{10mm}}{F} 
    & \multicolumn{1}{>{\centering\arraybackslash}m{10mm}}{A}
    & \multicolumn{1}{>{\centering\arraybackslash\columncolor{Gray}}m{9mm}}{F} 
    & \multicolumn{1}{>{\centering\arraybackslash}m{8mm}}{A}
    & \multicolumn{1}{>{\centering\arraybackslash}m{0mm}}{}\\
\midrule
1-2 & ${3.98}^{{\scriptscriptstyle C}}$ & ${4.50}^{{\scriptscriptstyle B}}$ & ${4.16}^{{\scriptscriptstyle B}}$ & ${4.61}^{{\scriptscriptstyle A}}$ & ${4.28}^{{\scriptscriptstyle A}}$ & ${4.66}^{{\scriptscriptstyle A}}$ &\\
$\geq3$ & ${3.91}^{{\scriptscriptstyle C}}$ & ${4.33}^{{\scriptscriptstyle B}}$ & ${4.03}^{{\scriptscriptstyle B}}$ & ${4.43}^{{\scriptscriptstyle B}}$ & ${4.17}^{{\scriptscriptstyle A}}$ & ${4.60}^{{\scriptscriptstyle A}}$& \\
\bottomrule
\end{tabular}}
\vspace{-1mm}
\caption{Fluency (F) and Adequacy (A) obtained in the human evaluation. \#T refers to the number of input triples and \#D to graph diameters. The ranking was determined by pair-wise Mann-Whitney tests with p < 0.05, and the difference between systems which have a letter in common is not statistically significant.}
\label{tab:humanevevaluation}
\end{table}


 \begin{figure*}[t]
    \centering
    \includegraphics[width=1\textwidth]{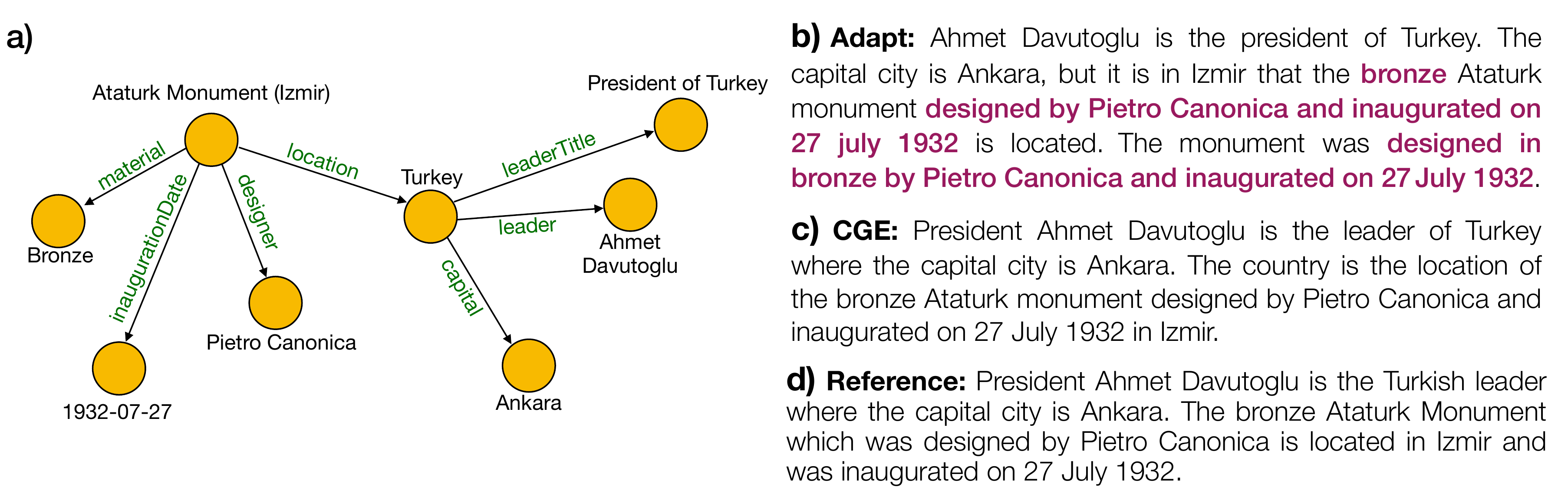}
    \vspace{-8mm}
    \caption{(a) A WebNLG input graph and the outputs for (b) Adapt and (c) CGE. The color face indicates repetition.}
    \label{fig:examples}
\end{figure*}

\subsection{Additional Experiments}

\paragraph{Impact of the Vocabulary Sharing and Length Penalty.}

During the ablation studies, we note that the vocabulary sharing and length penalty are beneficial for the performance. To better estimate their impact, we evaluate {\fontfamily{qcr}\selectfont CGE-LW} model with its variations without employing vocabulary sharing, length penalty and without both mechanisms, on the test set of both datasets. Table~\ref{tab:ablation-test} shows the results. We observe that sharing vocabulary is more important to WebNLG than AGENDA. This suggests that sharing vocabulary is beneficial when the training data is small, as in WebNLG. On the other hand, length penalty is more effective for AGENDA, as it has longer texts than WebNLG\footnote{As shown in Table~\ref{tab:datastatistics}, AGENDA has texts 5.8 times longer than WebNLG on average.}, improving the BLEU score by 0.71 points.

 \begin{figure}[t]
    \centering
    \includegraphics[width=.35\textwidth]{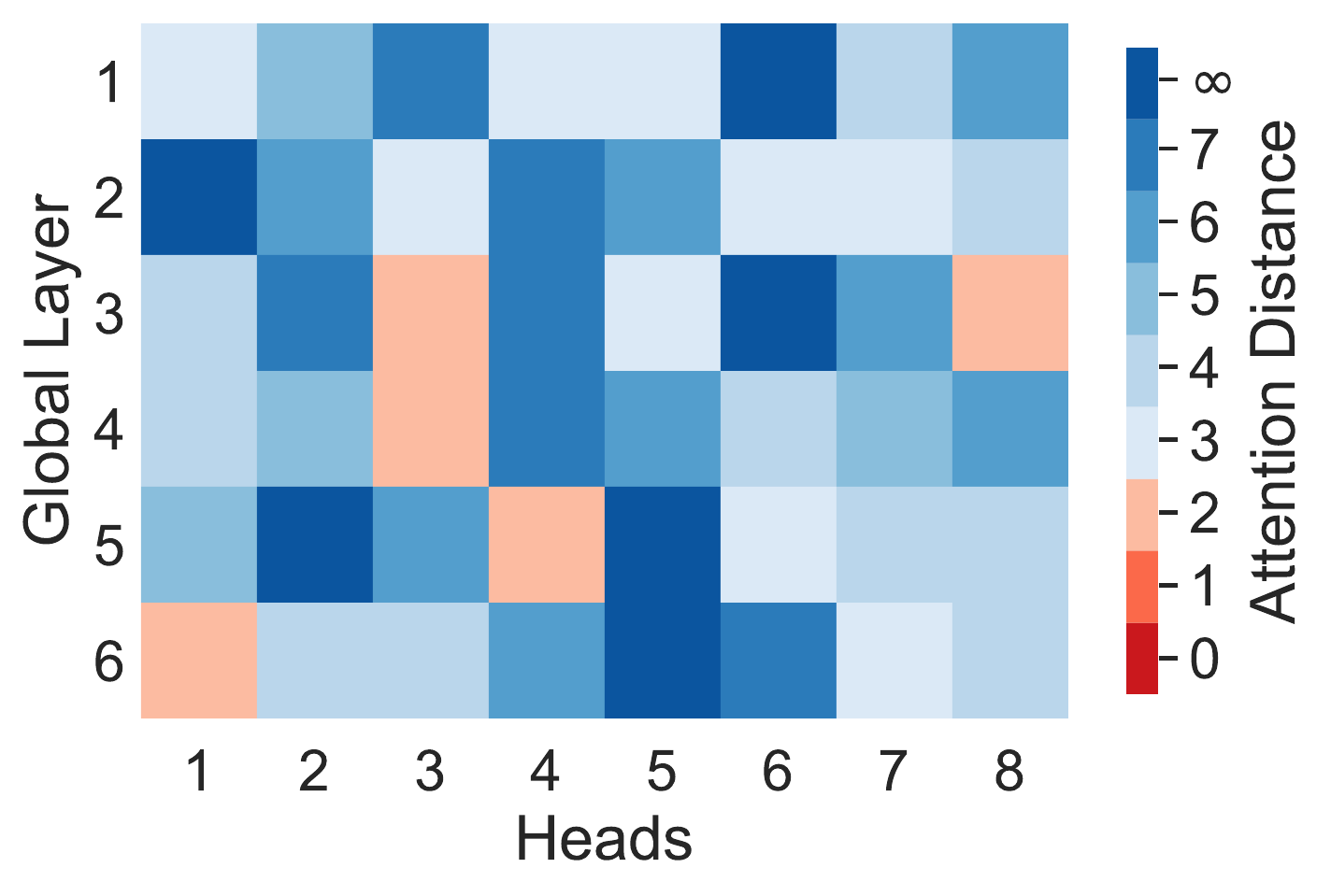}
    \vspace{-4mm}
    \caption{The average distance between nodes for the maximum attention for each head. $\infty$ indicates no path between two nodes, that is, they belong to distinct connected components.}
    \label{fig:attns}
    \vspace{-4mm}
\end{figure}
\paragraph{How Far Does the Global Attention Look At.}

Following previous works \cite{voita-etal-2019-analyzing, cai-lam-2020-graph}, we investigate the attention distribution of each graph encoder global layer of {\fontfamily{qcr}\selectfont CGE-LW} on the AGENDA dev set. In particular, for each node, we verify its global neighbor that receives the maximum attention weight and record the distance between them.\footnote{The distance between two nodes is defined as the number of edges in a shortest path connecting them.} Figure~\ref{fig:attns} shows the averaged distances for each global layer. We observe that the global encoder mainly focuses on distant nodes, instead of the neighbours and closest nodes. This is very interesting and agrees with our intuition: whereas the local encoder is concerned about the local neighborhood, the global encoder is focusing on the information from long-distance nodes.
\vspace{-2mm}
\paragraph{Case Study.}
Figure~\ref{fig:examples} shows examples of generated texts when the WebNLG graph is complex (7 triples). While {\fontfamily{qcr}\selectfont CGE} generates a factually correct text (it correctly verbalises all triples), the Adapt's output is repetitive. The example also illustrates how the text generated by {\fontfamily{qcr}\selectfont CGE} closely follows the graph structure whereby the first sentence verbalises the right-most subgraph, the second the left-most one and the linking node \textit{Turkey} makes the transition (using hyperonymy and a definite description, i.e., \textit{The country}). The text created by {\fontfamily{qcr}\selectfont CGE} is also more coherent than the reference. As noted above, the input graph includes two subgraphs linked by \textit{Turkey}. In natural language, such a meaning representation corresponds to a topic shift with the first part of the text describing an entity from one subgraph, the second part an entity from the other subgraph, and the linking entity (\textit{Turkey}) marking the topic shift. Typically, in English, a topic shift is marked by a definite noun phrase in the subject position. While this is precisely the discourse structure generated by {\fontfamily{qcr}\selectfont CGE} (\textit{Turkey} is realised in the second sentence by the definite description \textit{The country} in subject position), the reference fails to mark the topic shift, resulting in a text with weaker discourse coherence.

\section{Conclusion}

In this work, we introduced a unified graph attention network structure for investigating graph-to-text models that combine global and local graph encoders in order to improve text generation. An extensive evaluation of our models demonstrated that the global and local contexts are empirically complementary, and a combination can achieve state-of-the-art results on two datasets. In addition, cascaded architectures give better results compared to parallel ones. 

We point out some directions for future work. First, it is interesting to study different fusion strategies to assemble the global and local encodings; Second, a promising direction is incorporating pre-trained contextualized word embeddings in graphs; Third, as discussed in Section~\ref{sec:outputlength}, it is worth studying ways to diminish the gap between the reference and the generated text lengths. 

\section*{Acknowledgments}
We would like to thank Pedro Savarese, Markus Zopf, Mohsen Mesgar, Prasetya Ajie Utama, Ji-Ung Lee and Kevin Stowe for their feedback on this work, as well as the anonymous reviewers for detailed comments that improved this paper.
This work has been supported by the German Research Foundation as part of the Research Training Group Adaptive Preparation of Information from Heterogeneous Sources (AIPHES) under grant No. GRK 1994/1.

\bibliography{tacl2018}
\bibliographystyle{acl_natbib}

\end{document}